\documentclass[10pt,twocolumn,letterpaper]{article}

\usepackage{cvpr}              % To produce the CAMERA-READY version

\usepackage[table,xcdraw]{xcolor}
\usepackage{array}
\usepackage{tabularx}
\usepackage{arydshln}

\definecolor{cvprblue}{rgb}{0.21,0.49,0.74}
\usepackage[pagebackref,breaklinks,colorlinks,allcolors=cvprblue]{hyperref}

\definecolor{mycolor}{rgb}{0.749,0.361,0.820}

\def\confName{CVPR}
\def\confYear{2027}

\title{Efficient Latent Reasoning for Vision-Language Model \\via Dynamics-Aware Quantization and Distillation}

\author{Yanjing Li\\
Nanyang Technological University\\
\and
Sheng Xu\\
Communication University of China\\
\and
Baochang Zhang\\
Beihang University\\
\and
Xianbin Cao\\
Beihang University\\
\and
Yawei Li$^\dagger$\\
Nanyang Technological University\\
}

\begin{document}
\maketitle
\footnote{$^\dagger$: Corresponding author.}
\begin{abstract}
Efficient chain-of-thought (CoT) reasoning in vision-language models (VLMs) requires reducing both intermediate reasoning steps and their execution cost.
Latent CoT methods compress textual rationales into continuous thoughts, yet their full-precision (FP) models retain substantial storage and computation costs.
Lowering numerical precision could reduce these remaining costs, yet direct quantization severely degrades reasoning accuracy, particularly with 4-bit weights and activations.
Operator and perturbation analyses reveal key sensitivity and direction-dependent predictive effects of continuous-thought errors, motivating preservation of information retrieval and subsequent reasoning beyond reconstruction accuracy.
We therefore introduce \textbf{QL-VLM}, a dynamics-aware \textbf{Q}uantization framework for efficient \textbf{L}atent reasoning in Vision-Language Models (\textbf{VLM}s). 
First, Dynamics-Guided Quantization (DGQ) uses nonuniform codebooks that preserve rotary-key structure, mitigating quantization-induced distortion in vision-language information retrieval.
Second, Dynamics-Preserving Distillation (DPD) aligns continuous thoughts, latent-query readouts, and predictions with a full-precision teacher, maintaining predictive consistency through subsequent low-precision reasoning.
% 
% On ScienceQA-IMG with Qwen2-VL-2B-Instruct, 4-bit QL-VLM achieves 90.43\% reasoning accuracy and improves baseline by 40.33\%, with 16.64$\times$ lower theoretical inference cost than explicit CoT.
On ScienceQA-IMG with Qwen2-VL-2B-Instruct, 4-bit QL-VLM achieves 90.43\% accuracy and improves baseline by 40.33\%, with 16.64$\times$ fewer bit-operations and 6.40$\times$ lower measured latency than explicit CoT.

% \yawli{Abstract is a little bit long. Should compress by a few lines.}

% \yawli{People do not remember so many numbers. Mention a few that you want them to remember most.}

% \yawli{The abstract should raise readers' interest to continue. Introduction explains a little bit more. Think about how you can raise readers' interest.}

% 
% \yawli{There should be one highlighted number, used to sell the story.}
% \yawli{What does QL mean?}

\end{abstract}

\section{Introduction}

% \yawli{Are there actual acceleration results? for example, using 8-bit or 4-bit kernels?}

% \yawli{How long can a CoT be? How many tokens? Can the method be applied to very long CoT? How should the latent steps be adjusted?}
% Multimodal chain-of-thought (CoT) enables vision-language models (VLMs) to organize visual evidence and textual knowledge through intermediate rationales~\cite{zhang2023multimodal,xu2025llava}. However, generating these rationales requires sequential decoder updates and enlarges the attention history. On MMStar~\cite{chen2024we}, explicit CoT averages 142.1 reasoning tokens and incurs 7.24s of generation time, compared with 0.47s for input encoding, using Qwen3-VL-8B on a single H200 GPU~\cite{hu2026colt}. Longer reasoning traces further increase this burden, Kimi-VL-Thinking-2506 averages approximately 4.4K output tokens per example on MathVision, including reasoning and the final response~\cite{team2025kimi}. Each decoding update also executes a large model whose weights must remain accessible throughout inference. Efficient deployment therefore requires reducing both reasoning length and the storage and computation costs of the underlying model.

Multimodal chain-of-thought (CoT) enables vision-language models (VLMs) to integrate visual evidence and textual knowledge through intermediate rationales~\cite{zhang2023multimodal,xu2025llava}, at an inference cost that grows along two axes. The first is the \emph{number of decoding steps}: on MMStar~\cite{chen2024we}, explicit CoT with Qwen3-VL-8B spends 7.24\,s generating 142.1 reasoning tokens against 0.47\,s encoding the input~\cite{hu2026colt}, and thinking-style models further push output to several thousand tokens per example (about 4.4K for Kimi-VL-Thinking-2506)~\cite{team2025kimi}. The second is the \emph{cost per step}: each token or thought requires a forward pass through a multi-billion-parameter backbone whose weights, activations, and attention history are stored and read at full precision. Efficient deployment must reduce both.

\begin{figure}[t]
    \centering
    \includegraphics[scale=.64]{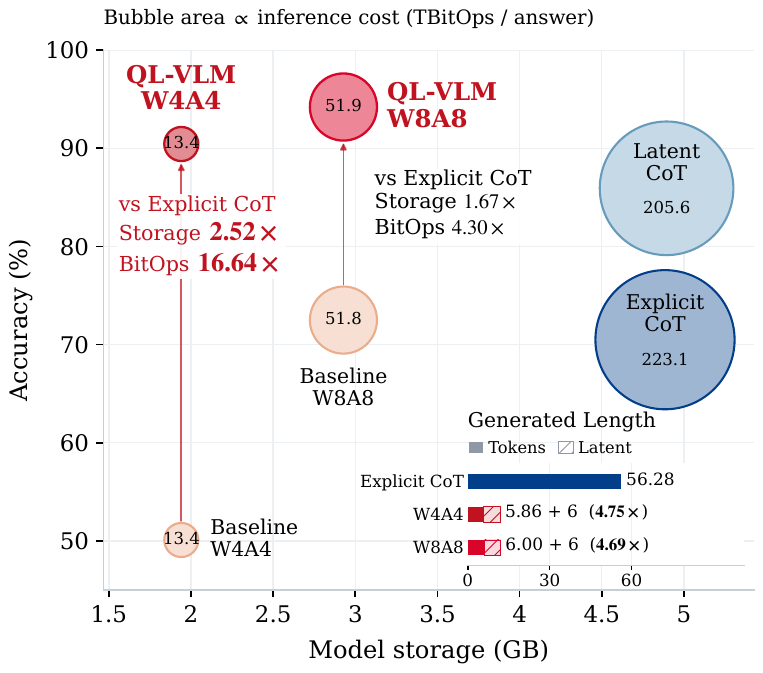}
    \caption{Accuracy-efficiency trade-off on ScienceQA-IMG with Qwen2-VL-2B-Instruct. 
    Bubble area denotes inference cost (TBitOps per answer), and arrows show gains over matched-precision baselines.   
    The inset compares generated tokens (solid) and latent steps (hatched).}
    % \yawli{Refrain from using bold and red color everywhere. highlight the ones that you really want people to remember.}
    % \yawli{Compression ratio should be annotated in the figure.}
    % Computation is BitOps-normalized.
    % \yawli{don't use png format. use pdf or svg or eps formats.} \yawli{W4A4 accuracy is still not on par with FP. This can be questioned. Useful to check this paper \href{https://arxiv.org/abs/2601.22709}{Gated Relational Alignment via Confidence-based Distillation for Efficient VLMs}}
    \label{fig:bubble}
\end{figure}

\begin{figure*}[htbp]
    \centering
    \includegraphics[scale=.52]{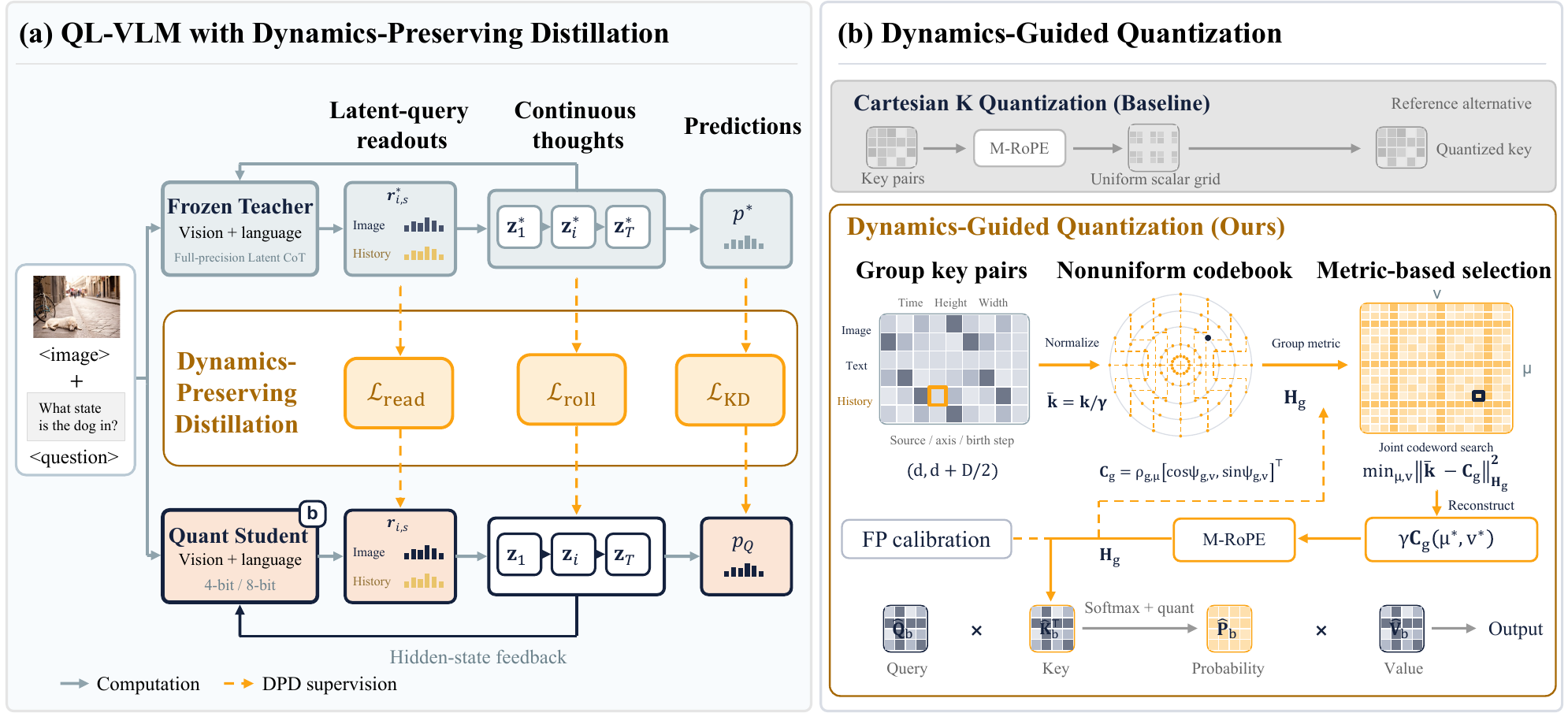}
    \caption{Overview of QL-VLM. (a) Dynamics-Preserving Distillation aligns the quantized student's continuous thoughts, latent-query readouts, and predictions with a frozen full-precision teacher. (b) Compared with Cartesian key quantization, Dynamics-Guided Quantization reconstructs rotary key pairs using grouped nonuniform codebooks and a calibration-based predictive metric at the same bit width. }
    \label{fig:framework}
\end{figure*}

% Prior work largely treats the two axes separately. Input-token pruning~\cite{chen2024image} and concise-reasoning objectives~\cite{xu2025chain} shorten the sequence, and latent CoT methods~\cite{hao2024training,shen2025codi,hu2026colt} replace textual rationales with a few continuous thoughts, cutting decoding length by an order of magnitude. Each remaining step, however, still executes a full-precision model. On ScienceQA~\cite{lu2022learn} with Qwen2-VL-2B (Fig.~\ref{fig:bubble}), full-precision latent CoT~\cite{shen2025codi} reduces the generated sequence from 56.28 tokens to six thoughts and six answer tokens, yet the model still occupies 4.89\,GB, and its theoretical per-answer compute, measured as MACs weighted by operand bit-width products, falls only from 223.10 to 205.65 TBitOps. Low-bit quantization is the established remedy for the second axis: post-training and quantization-aware methods~\cite{frantar2022gptq,xiao2023smoothquant,liu2024llm} reach 8-bit, and increasingly 4-bit, weights and activations on LLMs and VLMs with modest loss.

Prior work largely treats the two axes separately. Token pruning~\cite{chen2024image} and concise-reasoning objectives~\cite{xu2025chain} shorten the sequence, and latent CoT~\cite{hao2024training,shen2025codi,hu2026colt} goes further by replacing textual rationales with a few continuous thoughts; each remaining step, however, still executes a full-precision model. On ScienceQA~\cite{lu2022learn} with Qwen2-VL-2B (Fig.~\ref{fig:bubble}), full-precision latent CoT~\cite{shen2025codi} cuts generation from 56.28 tokens to six thoughts and six answer tokens, but the model still occupies 4.89\,GB and 
its theoretical per-answer compute falls by only 8\% (223.10 to 205.65 TBitOps).
% its theoretical per-answer compute (MACs weighted by operand bit-width products) falls only from 223.10 to 205.65 TBitOps. 
Low-bit quantization addresses the second axis: post-training and quantization-aware methods~\cite{frantar2022gptq,xiao2023smoothquant,liu2024llm} reach 8-bit, and increasingly 4-bit, weights and activations with modest loss.

Yet, the two remedies do not combine additively. Quantizing the latent-CoT model to 4-bit weights and activations lowers ScienceQA accuracy from 79.97\% to 11.16\%, and quantization-aware training recovers it only to 50.10\% (Fig.~\ref{fig:bubble}). We attribute this to \textbf{the structure of latent reasoning} rather than to tuning. In explicit CoT, a quantization error perturbs a single token that is immediately re-encoded as a discrete symbol. In latent CoT, continuous thoughts are fed back as the next input embeddings, while their cached keys remain accessible to subsequent queries. These pathways allow errors to propagate through later reasoning and affect answer predictions.

Two analyses support this account and determine our design. Operator ablations (Sec.~\ref{sec:diagnose}) identify attention keys as the dominant bottleneck: with linear layers at W4A4, raising only the keys from 4 to 8 bits lifts accuracy from 11.16\% to 70.35\%, against at most 33.41\% for queries, values, or attention probabilities. A perturbation analysis using the total Jacobian of the answer logits through subsequent latent steps further shows that predictive impact depends on direction, not only magnitude: equal-norm perturbations of continuous thoughts differ in induced KL by a median factor of 2.82, and a direction-aware estimate ranks them with Spearman correlation 0.999 versus 0.270 for a magnitude-based one (Fig.~\ref{fig:f2_local}). Reconstruction error, the objective of most quantizers, is therefore the wrong target; what must be preserved is the information later steps retrieve and the reasoning that consumes it.

We therefore introduce \textbf{QL-VLM}, a dynamics-aware \textbf{Q}uantization framework for low-bit \textbf{L}atent reasoning in \textbf{VLM}s (Fig.~\ref{fig:framework}), whose two components correspond to these two findings. \emph{Dynamics-Guided Quantization} (DGQ) quantizes each rotary key pair jointly in polar coordinates with a learned nonuniform codebook, so that the angular structure imposed by M-RoPE~\cite{su2024roformer,wang2024qwen2} survives quantization, and selects codewords under a prediction-sensitivity metric rather than Euclidean distance, with codebooks grouped by layer, head, rotary axis, key source, and the reasoning step at which a latent key is created. \emph{Dynamics-Preserving Distillation} (DPD) trains the quantized student against a full-precision teacher by aligning the information that latent queries retrieve from multimodal tokens and earlier thoughts, the continuous thoughts weighted by their estimated downstream sensitivity, and the answer distribution, so that the student's reasoning trajectory, not only its output, follows the teacher's. Both vision and language backbones are quantized.

On ScienceQA with Qwen2-VL-2B, QL-VLM reaches 90.43\% at 4 bits and 94.20\% at 8 bits, improving over the matched-precision baseline by 40.33 and 7.49 points; across three backbones and three datasets, the 4-bit gains range from 21.37 to 48.39 points, and the quantized models match or exceed full-precision latent CoT in eight of nine settings. The 4-bit model generates 5.86 answer tokens on average, shortening the decoding sequence by 4.75$\times$ relative to explicit CoT-SFT once the six thoughts are counted, with 2.52$\times$ lower storage. Measured on an RTX PRO 6000 GPU, it answers in 0.50\,s per question, against 3.20\,s for explicit CoT and 0.80\,s for full-precision latent CoT.

% \yawli{some fundamental questions}
% \yawli{1. Do six latent steps work for CoT of any length? I don't think so. Perhaps there is a sweet point for choosing the compression ratio rather than compressing to a limited number of steps.}
% \yawli{2. Should there be differences in treating the language tokens and vision tokens in the latent steps?}

\section{Related Work}
\subsection{Efficient Chain-of-Thought}
Efficient reasoning methods for Large-Language Models (LLMs) can be grouped into input-based, output-based, and model-based approaches~\cite{sui2025stop}. Input-based methods encourage concise reasoning through prompting, as in Chain of Draft~\cite{xu2025chain}, while prompt compression such as LLMLingua~\cite{jiang2023llmlingua} removes redundant input tokens. Output-based methods~\cite{deng2024explicit,xu2025softcot} replace textual traces with latent representations: Coconut~\cite{hao2024training} reuses hidden representations as inputs, CODI~\cite{shen2025codi} learns continuous thoughts through self-distillation, and SIM-CoT~\cite{wei2026sim} adds step-level supervision through a training-only decoder. Model-based methods optimize reasoning length through length-aware rewards or variable-length training data, exemplified by O1-Pruner~\cite{luo2026o1}.
For VLMs, efficient reasoning must additionally preserve visual evidence and cross-modal interactions. FastV~\cite{chen2024image} prunes visual inputs, while Skywork R1V~\cite{peng2025skywork} reduces excessive reasoning through adaptive-length distillation. Heima~\cite{shen2025efficient} compresses multimodal rationales into abstract thinking tokens, and CoLT~\cite{hu2026colt} learns compact latent thoughts through decoder-based and internal supervision. These methods primarily compress token sequences while retaining backbone precision, leaving substantial storage and arithmetic costs. How low-bit errors affect multimodal retrieval and subsequent latent reasoning remains insufficiently characterized.

\subsection{Vision Language Model Quantization}
Quantization methods broadly comprise post-training quantization (PTQ) and quantization-aware training (QAT). PTQ relies on calibration and reconstruction without end-to-end retraining. For LLMs, GPTQ~\cite{frantar2022gptq} uses second-order weight reconstruction, while SmoothQuant~\cite{xiao2023smoothquant} redistributes activation outliers. For VLMs, Q-VLM~\cite{wang2024q} considers cross-layer dependencies, MBQ~\cite{li2025mbq} balances modality-dependent sensitivity, and MQuant~\cite{yu2025mquant} introduces modality-specific scaling and outlier suppression. RGSQ~\cite{wu2026rgsq} further incorporates Fisher–Riemannian reconstruction geometry. 
KV-cache quantization additionally accounts for the distinct distributions of keys and values~\cite{liu2024kivi}. Polar-coordinate approaches~\cite{han2025polarquant,wu2026polarquant} further exploit recursively transform KV vectors before quantization.
Nevertheless, aggressive weight–activation quantization can substantially degrade accuracy, motivating training-based adaptation, particularly at lower bits.
QAT adapts model parameters under quantization during optimization. LLM-QAT~\cite{liu2024llm} uses generated data and distillation to quantize weights, activations, and KV caches, while NVFP4-QAD~\cite{xin2026quantization} recovers LLM and VLM accuracy by matching full-precision output distributions. These methods do not explicitly address how quantized keys alter information retrieval or how intermediate errors affect subsequent continuous thoughts, leaving their effectiveness for low-bit multimodal latent reasoning insufficiently established.

\section{Methodology}
QL-VLM combines low-bit quantization with multimodal latent reasoning, using a full-precision reference to supervise the compressed model. We introduce the preliminaries in Sec.\,\ref{sec:preliminary} and diagnose quantization sensitivity in Sec.\,\ref{sec:diagnose}. This analysis motivates Dynamics-Guided Rotary Key Quantization in Sec.\,\ref{sec:dgq} and Dynamics-Preserving Latent Reasoning Distillation in Sec.\,\ref{sec:dpd}.

\subsection{Preliminaries}\label{sec:preliminary}
\noindent\textbf{Vision-language latent reasoning}. 
We adapt CODI~\cite{shen2025codi} to a VLM, which jointly processes visual tokens and a textual question as the multimodal prompt $x$. Explicit-CoT and implicit-CoT branches share the same full-precision parameters $\theta$. The implicit branch replaces textual reasoning with $T$ continuous thoughts indexed by $i\in{1,\ldots,T}$, bounded by learnable \texttt{<bot>} (begin-of-thoughts) and \texttt{<eot>} (end-of-thoughts) tokens:
\begin{equation}
    \begin{aligned}
        f_0 &= F_\theta(x),\qquad Z_i=\Pi_\theta(f_{i-1}),\\
        f_i &= F_\theta(x, Z_{1:i}).
    \end{aligned}
\end{equation}
where $i\in\{1,\ldots,T\}$, $F_\theta$ returns the last decoder hidden vector, and $\Pi_\theta$ is a two-layer MLP followed by LayerNorm. Each $\mathbf z_i$ becomes the next input embedding. After $T$ thoughts, a learnable $\texttt{<eot>}$ token is manually inserted before answer decoding.
Both branches receive token-level cross-entropy supervision. CODI additionally aligns their hidden activations as:
\begin{equation}
    \mathcal L_{\mathrm{bd}}^{\mathrm{FP}} = \mathbb E_{l}\left[\frac{1}{\sigma_l}\mathbb E_{\mathrm{coord}}\left[|b_{l,\mathrm{coord}} - \operatorname{sg}(b_{l,\mathrm{coord}}^{\mathrm{exp}})|\right]\right],
\end{equation}
Here, $b_{\ell,\mathrm{coord}}$ and $b_{\ell,\mathrm{coord}}^{\mathrm{exp}}$ denote implicit and explicit hidden coordinates at layer $\ell$; $\sigma_\ell$ is the detached teacher-activation standard deviation within the batch. Both expectations are uniform empirical averages, and $\operatorname{sg}$ stops target gradients. 
The shared full-precision model is trained jointly:
\begin{equation}\label{eq:loss_codi}
    \mathcal L_{\mathrm{CODI}} = \mathcal L_{\mathrm{ans}} + \lambda_{\mathrm{cot}}\mathcal L_{\mathrm{cot}} + \lambda_{\mathrm{bd}}\mathcal L_{\mathrm{bd}}^{\mathrm{FP}},
\end{equation}
where $\mathcal L_{\mathrm{ans}}$ supervises the implicit answer, $\mathcal L_{\mathrm{cot}}$ supervises the explicit rationale and answer, and $\lambda_{\mathrm{cot}},\lambda_{\mathrm{bd}}\ge0$ are loss weights. Inference uses only the implicit branch. For subsequent QAT, we adapt the implicit answer loss by normalizing its segments separately:
\begin{equation}\label{eq:ans_adapt}
    \tilde{\mathcal L}_{\mathrm{ans}} = \mathcal L_{\mathcal O} + \lambda_{\mathrm p}\mathcal L_{\mathcal P} + \lambda_{\mathrm e}\mathcal L_{\mathcal E}.
\end{equation}
Here, $\mathcal O$, $\mathcal P$, and $\mathcal E$ contain valid shifted prediction positions for the option token sequence, answer template, and final stop token, respectively.
% Each $\mathcal L_{\mathcal B}$ is the cross-entropy averaged over the valid tokens in segment $\mathcal B$, excluding padding. 
% We set $\lambda_{\mathrm p}=\lambda_{\mathrm e}=0.1$.

\noindent\textbf{Quantization}. 
Quantization reconstructs a tensor $\mathbf X$ as
\begin{equation}\label{eq:uniform_quant}
    \mathcal Q_b(\mathbf X;\Delta,o) = \Delta[\operatorname{clip}(\lfloor \mathbf X/\Delta \rceil +o, q_{\min},q_{\max})-o],
\end{equation}
where $b$ is the bit width, $\Delta>0$ the scale, $o$ the integer zero point, and $q_{\min},q_{\max}$ the integer bounds. Attention consumes reconstructed operands:
\begin{equation}\label{eq:q_attn}
    \begin{aligned}
        \mathbf P &= \operatorname{softmax}(\hat{\mathbf Q}\hat{\mathbf K}^{\mathsf T}/\sqrt{D} + \mathbf M), \\
        \mathbf O &= \hat{\mathbf P}\hat{\mathbf V},
    \end{aligned}
\end{equation}
where $\mathbf Q,\mathbf K,\mathbf V$ are queries, keys, and values, $D$ is the head width, $\mathbf M$ the attention mask, and $\mathbf O$ the output. Hats indicate quantized reconstructions. Subsequent sections use $\mathrm Q$ and $\star$ for quantized and full-precision execution.
% probabilities use a logarithmic grid without subsequent renormalization. 

\subsection{Quantization Sensitivity in Latent CoT}\label{sec:diagnose}
\begin{figure}[htbp]
    \centering
    \includegraphics[scale=.43]{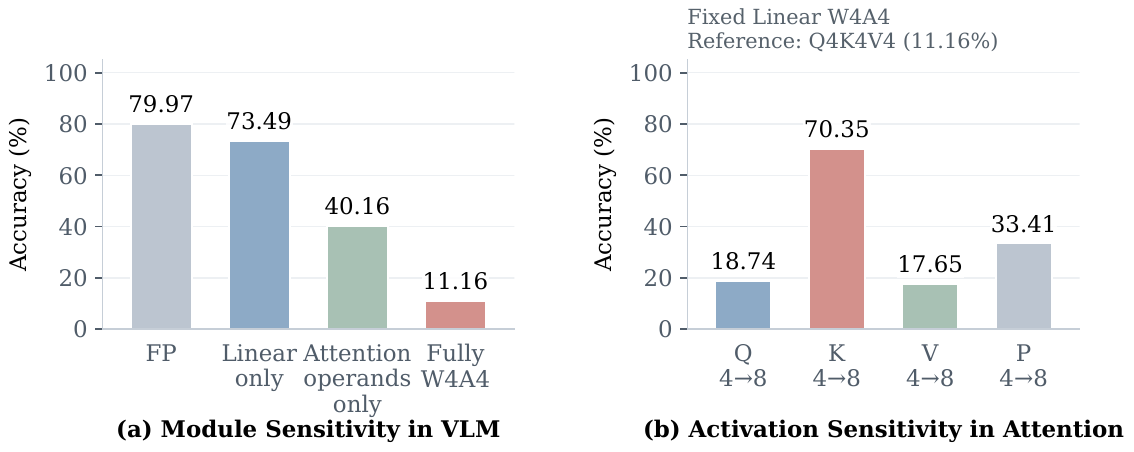}
    \caption{Quantization sensitivity of Qwen2-VL-2B-Instruct on ScienceQA. (a) Module-level comparison between FP and 4-bit Linear layers, attention operands, or both. (b) Activation-level comparison with 4-bit Linear layers: the precision of Q (query), K (key), V (value), or P (attention probability) is individually increased from 4 to 8 bits, while the remainder stays at 4 bits.}
    \label{fig:motivation_opt}
\end{figure}

\noindent\textbf{Operator sensitivity.} Fig.\,\ref{fig:motivation_opt}(a) shows that accuracy decreases from 79.97\% to 73.49\% with Linear-only quantization and 40.16\% with attention-operand-only quantization; quantizing both yields 11.16\%. With Linear layers fixed at W4A4, increasing only K from four to eight bits achieves 70.35\%, compared with 18.74\%, 17.65\%, and 33.41\% when increasing Q, V, and P, respectively (Fig.\,\ref{fig:motivation_opt}(b)). This identifies key precision as a major bottleneck under the tested protocol.
A possible contributor is cached-key reuse across reasoning steps, allowing key perturbations to affect multiple subsequent reads and continuous thoughts. This motivates using the answer logits to characterize perturbation effects and derive a prediction-sensitive error metric.

\begin{figure}[htbp]
    \centering
    \includegraphics[scale=.45]{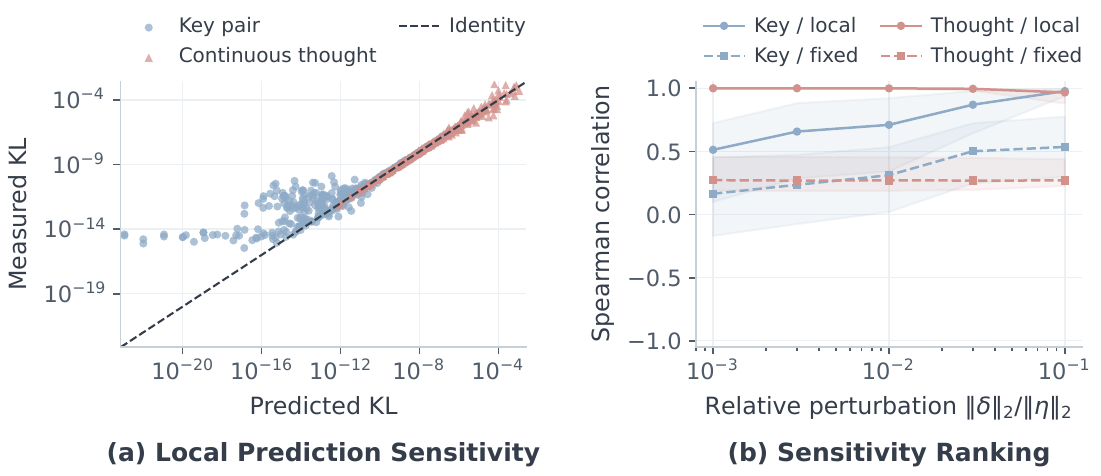}
    \caption{Prediction sensitivity of Qwen2-VL-2B trained with CODI on ScienceQA. (a) KL between unperturbed/perturbed predictions versus sample-specific quadratic estimates. (b) Spearman correlation with measured KL for local/fixed estimates, computed separately at each perturbation magnitude. Shading indicates 95\% bootstrap confidence intervals over 8 held-out examples.}
    \label{fig:f2_local}
\end{figure}

\noindent\textbf{Prediction sensitivity.}
We examine prediction sensitivity by perturbing one intermediate representation in full-precision implict CoT and recomputing subsequent latent reasoning, as illustrated in Fig.~\ref{fig:f2_local}. 
Let $\boldsymbol\eta$ denote a pre-M-RoPE key pair or a continuous thought, and $\boldsymbol\delta$ an additive perturbation. We measure the resulting change through the KL divergence between unperturbed and perturbed full-vocabulary prediction distributions~\cite{martens2020new}, using the same teacher-forced answer context. Under local smoothness, this change admits the second-order approximation as
\begin{equation}\label{eq:fisher_kl}
    \begin{aligned}
        &\sum_{n\in\mathcal N}D_{\mathrm{KL}}\!\left(\mathbf p_n^\star(\boldsymbol\eta)\,\middle\|\,\mathbf p_n^\star(\boldsymbol\eta+\boldsymbol\delta)\right) = \frac12\boldsymbol\delta^\top \mathcal F(\boldsymbol\eta)\boldsymbol\delta + o(\|\boldsymbol\delta\|_2^2),\\
        &\mathcal F(\boldsymbol\eta) = \sum_{n\in\mathcal N}\mathbf J_n^\top \left[\operatorname{diag}(\mathbf p_n^\star) - \mathbf p_n^\star\mathbf p_n^{\star\top}\right]\mathbf J_n,
    \end{aligned}
\end{equation}
where $\mathcal N$ contains the option and stop-token prediction positions, $\mathbf p_n^\star$ is the unperturbed reference distribution, and $\mathbf J_n=\partial\mathbf u_n^\star/\partial\boldsymbol\eta$ is the total Jacobian of the corresponding logits $\mathbf u_n^\star$, including paths through subsequent latent reasoning. The prediction Fisher metric $\mathcal F$ therefore captures local predictive effects beyond perturbation magnitude, detailed derivation and offline estimation are provided in the supplementary material.
Fig.~\ref{fig:f2_local} compares (a) measured/JVP-predicted KL and (b) local/fixed ranking agreement across perturbation magnitudes. At 1\% relative magnitude, nearly equal-norm perturbations across 48 thought sites from 8 held-out examples yield a median larger-to-smaller KL ratio of 2.82. Local predictions achieve a Spearman correlation of 0.999, versus 0.270 for the fixed approximation.

These results demonstrate direction-dependent predictive effects while exposing limitations of the tested fixed approximation. Together with the operator diagnosis, these findings motivate structure-preserving key quantization and full-precision teacher supervision of subsequent reasoning.

% \subsection{Dynamics-Guided Rotary Key Quantization}
\subsection{Dynamics-Guided Quantization}
\label{sec:dgq}
To preserve structured key representations for subsequent information retrieval, we use prediction sensitivity to guide rotary-pair reconstruction at a fixed bit width. Following Eq.\,(\ref{eq:fisher_kl}), we first group rotary key pairs and estimate a prediction-sensitive metric for each group, then use these metrics to guide codebook fitting and codeword selection.

\noindent\textbf{Rotary geometry and grouping.} Within a fixed language layer and KV head, let $\mathbf k_{j,d}$ be the raw key pair formed by channels $(d,d+D/2)$, where $j$ indexes a key token and $d\in\{1,\ldots,D/2\}$ a rotary pair. M-RoPE~\cite{wang2024qwen2}, most used in VLMs, assigns each pair to an axis $a\in\{\mathrm{temporal},\mathrm{height},\mathrm{width}\}$ and rotates it as
\begin{equation}
    \tilde{\mathbf k}_{j,d} = \mathbf R(\phi_{a,j,d})\mathbf k_{j,d},\qquad \mathbf R(\phi)= \begin{bmatrix}\cos\phi&-\sin\phi\\\sin\phi&\cos\phi\end{bmatrix},
\end{equation}
where $\phi_{a,j,d}$ is the angle supplied by the model's M-RoPE positions and frequencies. For fixed positions, the raw-coordinate metric satisfies
\begin{equation}
    \mathcal F(\mathbf k_{j,d}) = \mathbf R(\phi_{a,j,d})^\top \mathcal F(\tilde{\mathbf k}_{j,d}) \mathbf R(\phi_{a,j,d}).
\end{equation}
Although rotation preserves Euclidean length, the prediction metric need not be isotropic. We therefore preserve the rotary pairs and use their measured prediction sensitivity. Each codebook group is $g=(l,\kappa,a,\xi)$, where $l$ is the language layer, $\kappa$ the KV head, and $\xi\in\{\mathrm{text},\mathrm{image},\mathrm{latent}_1,\ldots,\mathrm{latent}_T\}$ the key source. A latent source records the step at which the key was created, independently of the later query that reads it.
For each group, we average sampled raw-pair Fisher blocks with sampling weights, add trace-scaled isotropic damping, and divide all groups by one common trace normalizer. The resulting positive metric is denoted $\mathbf H_g$. A common normalizer preserves relative sensitivity across groups.

\noindent\textbf{Polar codebooks.} Normalize a pair using $\gamma_j=\max_d\|\mathbf k_{j,d}\|_2$, floored at $10^{-8}$, within its token and KV head, and set $\bar{\mathbf k}_{j,d}=\mathbf k_{j,d}/\gamma_j$. Each group learns radius entries $\rho_{g,\mu}\in[0,1]$ and circular phase entries $\psi_{g,\nu}\in[-\pi,\pi)$, with indices $\mu,\nu\in\{0,\ldots,2^b-1\}$. One radius entry is fixed to zero. The normalized codeword is
\begin{equation}\label{eq:norm_codeword}
    \mathbf C_g(\mu,\nu) = \rho_{g,\mu}
    \begin{bmatrix}\cos\psi_{g,\nu}\\\sin\psi_{g,\nu}\end{bmatrix}.
\end{equation}
The selected indices $(\mu^\star,\nu^\star)$ minimize
\begin{equation}\label{eq:h_g}
    (\bar{\mathbf k}_{j,d}-\mathbf C_g(\mu,\nu))^{\mathsf T} \mathbf H_g (\bar{\mathbf k}_{j,d}-\mathbf C_g(\mu,\nu)).
\end{equation}
We reconstruct $\hat{\mathbf k}_{j,d}=\gamma_j\mathbf C_g(\mu^\star,\nu^\star)$ and apply the actual M-RoPE rotation to obtain the key operand in Eq.\,(\ref{eq:q_attn}). The factor $\gamma_j^2$ multiplies all candidate distortions equally for a given pair and therefore cancels in assignment. It is retained when fitting shared codebooks across pairs with different scales. Calibration alternates codeword assignment, closed-form radius updates, and periodic phase optimization. 
During QAT, we compute the key reconstruction loss by uniformly averaging over valid rotary key pairs within each quantizer invocation and then over active invocations:
\begin{equation}\label{eq:loss_key}
    \mathcal L_{\mathrm{key}}=\frac12\mathbb E_{\mathrm{call}}[\mathbb E_{\mathbf k\mid\mathrm{call}} [(\hat{\mathbf k}-\mathbf k)^\top\mathbf H_g(\hat{\mathbf k}-\mathbf k)]].
\end{equation}
Here, an invocation denotes one application of a layer’s key quantizer during a forward pass, $\mathbf k$ and $\hat{\mathbf k}$ denote the raw and reconstructed key pairs, and $g$ identifies their codebook group. The group metric $\mathbf H_g$ remains fixed and is also used for codeword selection at inference, without requiring gradients from future queries.
The main configuration fixes both indices to $b=4$: a pair uses eight logical index bits, or four per original key component. It does not assign higher precision to selected groups. We perform a chunked joint search over the $16\times16$ codewords on 4-bit runs. 8-bit runs use coordinate search. 
% Polar keys are used in the student's implicit reasoning and answer branch; the explicit-CoT branch and vision attention retain Cartesian key quantization.

This component incorporates reasoning dynamics into the key encoding rule. Quantized queries, probabilities, values, and subsequent parameter updates also affect the generated reasoning sequence, motivating supervision of the sequence and its retrieved information.

% \subsection{\color{red} Dynamics-Preserving Latent Reasoning Distillation}
\subsection{Dynamics-Preserving Distillation}
\label{sec:dpd}
To preserve how retrieved information supports subsequent latent reasoning, Dynamics-Preserving Distillation (DPD) aligns continuous thoughts, latent-query readouts, and answer distributions with a full-precision teacher parameterized by $\theta^\star$. The teacher is trained from the full-precision latent CoT~\cite{shen2025codi}, including its trained adapters and reasoning projection, and remains frozen in evaluation mode throughout training.
Let $\mathbf z_i$ and $\mathbf z_i^\star$ denote student and teacher thoughts at step $i\in\{1,\ldots,T\}$, each following its own reasoning history, where $T$ is the reasoning budget.
Using the diagonal calibration estimate $\mathbf G_i$ of $\mathcal F(\mathbf z_i^\star)$ in Eq.~(\ref{eq:fisher_kl}), we define the common normalizer $Z_F=\max(T^{-1}\sum_{i=1}^{T}\operatorname{tr}(\mathbf G_i),10^{-20})$.

\noindent\textbf{Latent-query readout alignment.}
We first align the information retrieved by latent queries from visual tokens and historical thoughts. For language layer $l$, query head $h$, reasoning step $i$, and source $s\in\{\mathrm{image},\mathrm{history}\}$, the student and teacher readouts are
\begin{equation}
    \begin{aligned}
        \mathbf r_{l,h,i,s} &= \sum_{j\in\mathcal J_{i,s}}\hat P_{l,h,i,j}\hat{\mathbf V}_{l,h,j}, \\
        \mathbf r^\star_{l,h,i,s} &= \sum_{j\in\mathcal J_{i,s}} P^\star_{l,h,i,j}\mathbf V^\star_{l,h,j}.
    \end{aligned}
\end{equation}
% Here, $\mathcal J_{i,\mathrm{image}}$ contains visible image-token positions, while $\mathcal J_{i,\mathrm{history}}$ contains preceding continuous thoughts, excluding the current thought. Softmax is computed over the full visible history before quantization and source selection, without source-wise renormalization. The readout therefore retains both the source's attention contribution and its value content.
Here, $\mathcal J_{i,\mathrm{image}}$ and $\mathcal J_{i,\mathrm{history}}$ contain visible image tokens and earlier thoughts, respectively. Softmax precedes probability quantization and source selection over the full visible history, without source-wise renormalization applied.
For each example, let $\mathcal A$ contain the valid selected layer--head--step--source tuples. We minimize
\begin{equation}\label{eq:loss_read}
    \begin{aligned}
        w_i &= \frac{\operatorname{tr}(\mathbf G_i)}{Z_F}, \\
        \mathcal L_{\mathrm{read}} &= \frac1{|\mathcal A|}\sum_{(l,h,i,s)\in\mathcal A}w_i\,\ell_1 (\frac{\mathbf r_{l,h,i,s}-\mathbf r^\star_{l,h,i,s}}{\sqrt{\max(\|\mathbf r^\star_{l,h,i,s}\|_2^2/D,\epsilon)}}),
    \end{aligned}
\end{equation}
where $\ell_1$ is the coordinate-averaged Smooth-L1 penalty, and $\epsilon=10^{-4}$ floors the reference mean-square norm. 
% We use 4 evenly spaced language layers, 2 query heads, and the last two thoughts, replaying only selected query rows with weights normalized over all $T$ thoughts.

\noindent\textbf{Sensitivity-weighted thought alignment.}
Readout alignment constrains the information accessed during reasoning; thought alignment further regularizes the intermediate representations supporting subsequent predictions. We use
\begin{equation}\label{eq:loss_roll}
    \mathcal L_{\mathrm{roll}} = \frac{1}{2T Z_F}\sum_{i=1}^T(\mathbf z_i-\mathbf z_i^\star)^\top\mathbf G_i(\mathbf z_i-\mathbf z_i^\star).
\end{equation}
Because each model follows its own reasoning history, the discrepancy includes the effects of earlier reasoning differences. The diagonal metrics weight thought coordinates according to their estimated downstream predictive sensitivity. As fixed local approximations, they provide sensitivity-based regularization without guaranteeing a monotonic reduction in final prediction KL.

\noindent\textbf{Answer-level supervision.}
We supervise answer correctness using the adapted loss $\tilde{\mathcal L}_{\mathrm{ans}}$ in Eq.~(\ref{eq:ans_adapt}) and preserve predictive behavior by distilling the frozen teacher's
full-vocabulary distributions at option positions:
\begin{equation}
    \mathcal L_{\mathrm{KD}} = \frac{\tau^2}{|\mathcal O|} \sum_{n\in\mathcal O} D_{\mathrm{KL}}[\operatorname{sg}(\operatorname{softmax}(\frac{\mathbf u_n^\star}{\tau})) \| \operatorname{softmax}(\frac{\mathbf u_n^{\mathrm Q}}{\tau})],
\end{equation}
where $\mathbf u_n^\star$ and $\mathbf u_n^{\mathrm Q}$ are teacher and student logits, $\operatorname{sg}$ stops gradients, and $\tau>0$ is the temperature~\cite{hinton2015distilling}, set to one by default. Each model conditions on its own latent reasoning history and the same preceding teacher-forced answer tokens. Ground-truth supervision remains active.

% We also retain the quantized student's explicit-CoT token CE, denoted $\mathcal L_{\mathrm{cot}}$, and CODI boundary alignment. Let $\mathbf b_l$ be the student implicit answer-boundary hidden vector and $\mathbf b_l^\star$ the frozen model's explicit-CoT boundary vector at layer $l$. Boundary supervision is
% \begin{equation}
%     \mathcal L_{\mathrm{bd}} = \operatorname{mean}_{l}\frac{\ell_1(\mathbf b_l-\mathbf b_l^\star)}{\max(\operatorname{std}(\mathbf b_l^\star),10^{-6})},
% \end{equation}
% where the standard deviation is computed over the reference vector's coordinates. Explicit rationales are used only in this training branch and its boundary target; the teacher's latent thoughts, readouts, and answer predictions are generated without conditioning on explicit CoT.

We retain the student's explicit-CoT CE $\mathcal L_{\mathrm{cot}}$ and boundary alignment $\mathcal L_{\mathrm{bd}}$, using a reference-standard-deviation-normalized Smooth-L1 penalty against the frozen teacher's explicit-CoT boundary activations. Explicit rationales are used only for this training branch and its boundary targets, not for the teacher's latent reasoning.

% Together, DGQ preserves structured key representations, while DPD supervises their information reads, subsequent continuous reasoning, and resulting predictions. The full-precision teacher is discarded at inference.

\noindent\textbf{Overall objective.} 
We combine answer-level supervision and CODI~\cite{shen2025codi} training losses in Eq.\,\ref{eq:loss_codi} with key reconstruction, thought alignment, and readout alignment:
\begin{equation}\label{eq:loss_all}
% \small
    \begin{aligned}
        \mathcal L=&
        \mathcal L_{\mathrm{CODI}} + \lambda_{\mathrm{KD}}\mathcal L_{\mathrm{KD}}
        \\
        &+\alpha_m\Big[
        \lambda_{\mathrm{key}}\mathcal L_{\mathrm{key}}
        +\chi M_{\mathrm{ref}}\big(
        \lambda_{\mathrm{read}}\mathcal L_{\mathrm{read}}
        +\lambda_{\mathrm{roll}}\mathcal L_{\mathrm{roll}}
        \big)\Big].
    \end{aligned}
\end{equation}
The first three terms follow CODI's implicit-answer supervision, explicit-CoT supervision, and answer-boundary alignment, with the adaptations described above. The remaining terms introduce frozen-teacher prediction distillation and dynamics-aware reconstruction and alignment.
All $\lambda$ are nonnegative loss weights. The factor $\alpha_m\in[0,1]$ gradually activates auxiliary supervision at optimizer update $m\ge0$. To reduce training overhead, readout and thought alignment are evaluated on micro-batches selected by $\chi\sim\operatorname{Bernoulli}(1/M_{\mathrm{ref}})$, where $M_{\mathrm{ref}}\ge1$ is an integer. Multiplication by $M_{\mathrm{ref}}$ preserves their expected loss contribution. Details are in Appendix.

% The complete training objective is
% \begin{equation}\label{eq:loss_all}
%     \begin{aligned}
%         \mathcal L={}&\mathcal L_{\mathrm{ans}} + \lambda_{\mathrm{KD}}\mathcal L_{\mathrm{KD}} + \lambda_{\mathrm{cot}}\mathcal L_{\mathrm{cot}} + \lambda_{\mathrm{bd}}\mathcal L_{\mathrm{bd}}\\
%         &+ \alpha_m[\lambda_{\mathrm{key}}\mathcal L_{\mathrm{key}} + \chi M_{\mathrm{ref}}\big(\lambda_{\mathrm{roll}}\mathcal L_{\mathrm{roll}} + \lambda_{\mathrm{read}}\mathcal L_{\mathrm{read}}\big)].
%     \end{aligned}
% \end{equation}
% Here, all $\lambda$ coefficients are nonnegative loss weights, the nonnegative integer $m$ indexes optimizer updates, and $\alpha_m\in[0,1]$ ramps the auxiliary losses. The indicator $\chi\in\{0,1\}$ selects rollout/readout micro-batches with probability $1/M_{\mathrm{ref}}$ when auxiliary supervision is active; the integer $M_{\mathrm{ref}}\ge1$ supplies the inverse-probability correction. Coefficients, schedules, and reference caching are specified in Sec.\,\ref{sec:exp_set}.

\begin{table}[t]
\centering
\small
\caption{Generated-answer accuracy (\%) on supervised vision-language CoT reasoning. Each dataset uses its own FP-trained initialization (gray-shadowed rows). 
Explicit CoT denotes CoT-SFT on ScienceQA-IMG and A-OKVQA, and Multimodal-CoT on M$^3$CoT. 
\#B specifies the corresponding bit width defined in Sec.\,\ref{sec:exp_set}. The best results are in \textbf{bold}. ``-'' indicates failure to converge under the specified training budget. }
\label{tab:main_reasoning}
\begin{tabular}{ccccc}
\toprule
Method & \#B & ScienceQA & A-OKVQA & M$^3$CoT \\
\hline
\multicolumn{5}{@{}c}{\textbf{Qwen2-VL-2B-Instruct}} \\ \hline
Explicit CoT                        & 16 & 70.50 & 71.32 & 42.52 \\
\rowcolor[HTML]{EFEFEF}
Latent CoT                          & 16  & 79.97 & 79.91 & 44.16 \\ \hline
Q-VLM~\cite{wang2024q}              & 4   & -     & -     & - \\
LLM-QAT~\cite{liu2024llm}           & 4   & -     & -     & - \\
Baseline                            & 4   & 50.10 & 34.10 & 20.40 \\
\textbf{QL-VLM}                     & 4   & \textbf{90.43} & \textbf{82.49} & \textbf{41.77} \\ \hdashline
Q-VLM~\cite{wang2024q}              & 8   & 72.50 & 70.35 & 31.39 \\
LLM-QAT~\cite{liu2024llm}           & 8   & 79.31 & 77.78 & 32.25 \\
Baseline                            & 8   & 86.71 & 83.59 & 34.50 \\
\textbf{QL-VLM}                     & 8   & \textbf{94.20} & \textbf{89.73} & \textbf{44.34} \\
\hline
\multicolumn{5}{@{}c}{\textbf{Qwen2.5-VL-7B-Instruct}} \\ \hline
Explicit CoT                        & 16 & 89.74 & 84.38 & 70.25 \\
\rowcolor[HTML]{EFEFEF}
Latent CoT                          & 16 & 90.74 & 88.89 & 71.10 \\ \hline
Q-VLM~\cite{wang2024q}              & 4   & 35.14 & 30.01 & - \\
LLM-QAT~\cite{liu2024llm}           & 4   & 40.26 & 38.72 & - \\
Baseline                            & 4   & 63.31 & 59.46 & 40.12 \\
\textbf{QL-VLM}                     & 4   & \textbf{93.78} & \textbf{89.72} & \textbf{71.34} \\ \hdashline
Q-VLM~\cite{wang2024q}              & 8   & 81.93 & 72.64 & 59.97 \\
LLM-QAT~\cite{liu2024llm}           & 8   & 86.67 & 79.68 & 61.42 \\
Baseline                            & 8   & 91.13 & 82.34 & 68.67 \\
\textbf{QL-VLM}                     & 8   & \textbf{96.25} & \textbf{91.22} & \textbf{75.69} \\ 
\hline
\multicolumn{5}{@{}c}{\textbf{InternVL2.5-8B}} \\ \hline
Explicit CoT                        & 16 & 97.68 & 85.59 & 75.70 \\
\rowcolor[HTML]{EFEFEF}
Latent CoT                          & 16 & 98.07 & 87.42 & 77.91 \\ \hline
Q-VLM~\cite{wang2024q}              & 4   & 61.58 & 40.92 & - \\
LLM-QAT~\cite{liu2024llm}           & 4   & 67.99 & 45.82 & - \\
Baseline                            & 4   & 70.92 & 60.97 & 44.32 \\
\textbf{QL-VLM}                     & 4   & \textbf{98.02} & \textbf{89.01} & \textbf{77.89} \\ \hdashline
Q-VLM~\cite{wang2024q}              & 8   & 69.26 & 44.27 & 63.20 \\
LLM-QAT~\cite{liu2024llm}           & 8   & 72.21 & 53.82 & 65.21 \\
Baseline                            & 8   & 79.89 & 69.80 & 69.87 \\
\textbf{QL-VLM}                     & 8   & \textbf{98.87} & \textbf{91.08} & \textbf{79.92} \\ 
\bottomrule
\end{tabular}
\end{table}

\section{Experiments}
\subsection{Experimental Setup}
\label{sec:exp_set}
\noindent\textbf{Datasets and models.}
We evaluate Qwen2-VL-2B-Instruct~\cite{wang2024qwen2}, Qwen2.5-VL-7B-Instruct~\cite{wu2025qwen}, and InternVL2.5-8B~\cite{chen2024expanding} on ScienceQA-IMG~\cite{lu2022learn} and A-OKVQA~\cite{schwenk2022okvqa}, and M$^3$CoT~\cite{chen2024m3cot}. Explicit-CoT references use CoT-SFT on the first two datasets and Multimodal-CoT on M$^3$CoT; latent-CoT references use a multimodal adaptation of CODI~\cite{shen2025codi}. We compare QL-VLM with Q-VLM, LLM-QAT, and a Cartesian QAT baseline. For zero-shot transfer, fixed ScienceQA-IMG-trained checkpoints are evaluated on MMStar~\cite{chen2024we} and MMMU-MC~\cite{yue2024mmmu} without target-data fine-tuning, recalibration, or model selection.

\noindent\textbf{Quantization and training.}
We quantize linear-layer weights and input activations, as well as attention queries, keys, values, and scores, to 4 or 8 bits in both vision and language backbones. The output head and reasoning projection remain in floating point. All QAT variants start from the same dataset-specific full-precision CODI checkpoint and use an independent frozen teacher. The Cartesian baseline shares the remaining quantizers and adapted CODI supervision as in Eq.\,(\ref{eq:loss_codi}). All models use six continuous thoughts and one QAT epoch. Details are provided in the appendix.

\subsection{Main Results}
\label{sec:main_results}
\noindent\textbf{CoT reasoning.}
Tab.\,\ref{tab:main_reasoning} shows that QL-VLM enables low-bit latent reasoning while achieving accuracy competitive with or superior to full-precision CoT across diverse backbones and datasets. 
First, QL-VLM preserves strong performance relative to both full-precision explicit and latent CoT. At 8-bit, it surpasses both counterparts across all nine settings. Even at 4-bit, it achieves 90.43\% on ScienceQA-IMG with Qwen2-VL-2B-Instruct, exceeding full-precision latent CoT by 10.46\%. On the more challenging M$^3$CoT, its 4-bit performance also closely matches or exceeds full-precision latent CoT with both larger backbones. 
Second, QL-VLM consistently outperforms conventional quantization methods at matched precision, improving over the 4-bit Cartesian baseline by 21.37–48.39 percentage points. The largest gain occurs on A-OKVQA with the 2B backbone, where accuracy increases from 34.10\% to 82.49\%. Moreover, QL-VLM converges in every evaluated setting, including those where Q-VLM and LLM-QAT fail to converge under the specified training budget. 
Notably, latent CoT outperforms explicit CoT, possibly by retaining task-relevant information without reproducing lengthy rationales, consistent with CODI~\cite{shen2025codi}.

\begin{table}[htbp]
\centering
\small
\caption{Zero-shot accuracy (\%) with a fixed ScienceQA-IMG-trained source checkpoint. \#B specifies the corresponding bit width defined in Sec.\,\ref{sec:exp_set}. The best results are in \textbf{bold}.}
\label{tab:zero_shot}
\begin{tabular}{
    >{\centering\arraybackslash}p{2cm}
    c
    >{\centering\arraybackslash}p{2cm}
    >{\centering\arraybackslash}p{2cm}
}
% \begin{tabularx}{\linewidth}{Xccc}
\toprule
Method & \#B & MMStar & MMMU-MC \\
\hline
\multicolumn{4}{@{}c}{\textbf{Qwen2-VL-2B-Instruct}} \\ \hline
Explicit CoT          & 16 & 40.50 & 31.20 \\
Latent CoT            & 16 & 43.28 & 32.24 \\ \hline
Baseline              & 4  & 30.11 & 24.19 \\
\textbf{QL-VLM}       & 4  & \textbf{44.27} & \textbf{31.18} \\ 
Baseline              & 8  & 37.26 & 28.79 \\
\textbf{QL-VLM}       & 8  & \textbf{45.67} & \textbf{33.10} \\
\hline
\multicolumn{4}{@{}c}{\textbf{Qwen2.5-VL-7B-Instruct}} \\ \hline
Explicit CoT          & 16 & 60.80 & 55.45 \\
Latent CoT            & 16 & 61.60 & 56.97 \\ \hline
Baseline              & 4  & 41.91 & 36.72 \\
\textbf{QL-VLM}       & 4  & \textbf{61.32} & \textbf{56.60} \\ 
Baseline              & 8  & 58.90 & 52.19 \\
\textbf{QL-VLM}       & 8  & \textbf{62.20} & \textbf{57.20} \\
\bottomrule
\end{tabular}
\end{table}

\noindent\textbf{Zero-shot generalization.}
Tab.\,\ref{tab:zero_shot} evaluates transfer from fixed ScienceQA-IMG-trained checkpoints to MMStar~\cite{chen2024we} and MMMU-MC~\cite{yue2024mmmu} without target-dataset adaptation. QL-VLM consistently outperforms baseline across both models and benchmarks. At 4-bit, it improves accuracy by 14.16\% and 6.99\% on MMStar and MMMU-MC with Qwen2-VL-2B-Instruct and by 19.41\%/19.88\% with Qwen2.5-VL-7B-Instruct. 
% The latter recovers performance to within 0.37\% of FP latent CoT on both benchmarks. 
At 8-bit, QL-VLM surpasses FP latent CoT across all four comparisons. These results indicate that its improvements extend beyond the training dataset, supporting the preservation of transferable multimodal reasoning under low-bit computation.

\begin{figure}[htbp]
    \centering
    \includegraphics[scale=.48]{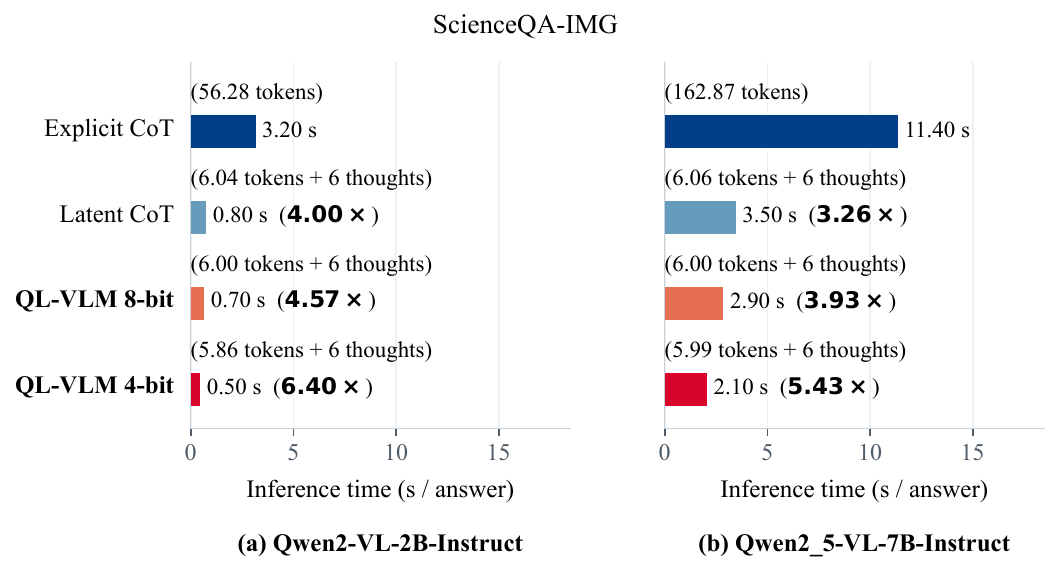}
    \caption{Inference efficiency on ScienceQA-IMG with (a) Qwen2-VL-2B-Instruct and (b) Qwen2.5-VL-7B-Instruct, measured on an NVIDIA RTX PRO 6000 GPU. Bars show inference time; annotations report generated tokens and continuous thoughts. Speedups are relative to explicit CoT within each model.}
    \label{fig:efficiency}
\end{figure}

\noindent\textbf{Efficiency and reasoning compression.}
Fig.\,\ref{fig:efficiency} compares inference time and decoding length on ScienceQA-IMG using an NVIDIA RTX PRO 6000 GPU. Supported low-bit matrix multiplications use compiled CUDA kernels, while the remaining use BF16 GEMM. 4-bit QL-VLM reduces average generated tokens from 56.28 to 5.86 with Qwen2-VL-2B-Instruct and from 162.87 to 5.99 with Qwen2.5-VL-7B-Instruct. Including six continuous thoughts, the decoding sequences are 4.75$\times$ and 13.58$\times$ shorter, respectively. 
Inference time decreases from 3.20 to 0.50 seconds per answer for the 2B model and from 11.40 to 2.10 seconds for the 7B model, yielding 6.40$\times$ and 5.43$\times$ speedups over explicit CoT. Compared with full-precision latent CoT, 4-bit QL-VLM provides additional 1.60$\times$ and 1.67$\times$ speedups; 8-bit QL-VLM also reduces latency on both models. These results demonstrate complementary reductions in reasoning length and measured execution time.

\subsection{Ablation studies}

\noindent\textbf{Component analysis.}
Tab.\,\ref{tab:dgq_progressive} progressively constructs DGQ
under fixed DPD supervision. Polar encoding and learned nonuniform codebooks improve accuracy over Cartesian keys, while axis-, source-, and birth-step-specific grouping provide further gains.
In particular, distinguishing historical thoughts by birth step increases accuracy from 72.69\% to 80.68\%, supporting reasoning-step-specific key reconstruction. Applying the predictive metric and key reconstruction loss further improves accuracy to 90.43\%. 
% These increments reflect contributions under the cumulative configuration.

\begin{table}[htbp]
\centering
\small
\caption{Progressive DGQ analysis on ScienceQA-IMG at 4-bit Qwen2-VL-2B-Instruct.
After enabling DPD, each row cumulatively adds the indicated design choice. Acc. denotes accuracy.}
\label{tab:dgq_progressive}
\begin{tabular}{llc}
\toprule
Configuration & Correspondence & Acc. (\%) \\
\midrule
Cartesian baseline
& Eqs.\,(\ref{eq:uniform_quant})-(\ref{eq:q_attn}) & 50.10 \\
\midrule
Cartesian K + DPD
& Eqs.\,(\ref{eq:loss_read})-(\ref{eq:loss_roll}) & 52.69  \\
\quad + Polar encoding
& Eq.\,(\ref{eq:norm_codeword}), fixed grids & 60.20  \\
\quad + Nonuniform codebooks
& Learn $\rho_{g,\mu},\psi_{g,\nu}$ & 64.28  \\
\quad + Axis grouping
& Axis index $a$ in $g$ & 70.11  \\
\quad + Source grouping
& Source index $\xi$ in $g$ & 72.69  \\
\quad + Birth-step grouping
& $\xi=\mathrm{latent}_i$ & 80.68  \\
\quad + Predictive metric
& $\mathbf H_g$ in Eq.\,(\ref{eq:h_g}) & 83.97  \\ \hdashline
\quad + $\mathcal L_{\mathrm{key}}$ (\textbf{QL-VLM})
& Eq.\,(\ref{eq:loss_key}) & \textbf{90.43}  \\
\bottomrule
\end{tabular}
\end{table}

Tab.\,\ref{tab:dpd_loss_ablation} separates the effects of thought and readout supervision with DGQ fixed. Both objectives improve over DGQ alone, with readout alignment providing larger gains on both backbones. Their combination reaches 90.43\% and 93.78\%, exceeding the stronger single-loss configuration by 17.69\% and 11.65\%, respectively. These results support the complementarity of aligning retrieved information and the continuous thoughts that support subsequent predictions.

\begin{table}[htbp]
\centering
\small
\caption{DPD ablations on ScienceQA-IMG at 4-bit setting. DGQ is enabled except in the Cartesian baseline. 2B (Qwen2-VL-2B-Instruct) and 7B (Qwen2.5-VL-7B-Instruct) report accuracy (\%).}
\label{tab:dpd_loss_ablation}
\begin{tabular}{lccccc}
\toprule
Variant &
$\lambda_{\mathrm{roll}}$ &
$\lambda_{\mathrm{read}}$ &
$\lambda_{\mathrm{KD}}$ &
2B & 7B \\
\midrule
Cartesian baseline
& 0 & 0 & 1 & 50.10 & 63.31 \\
\midrule
DGQ only
& 0 & 0 & 1 & 55.89 & 68.05 \\
$+\mathcal L_{\mathrm{roll}}$
& 0.05 & 0 & 1 & 69.72 & 79.79 \\
$+\mathcal L_{\mathrm{read}}$
& 0 & 0.05 & 1 & 72.74 & 82.13 \\ \hdashline
\begin{tabular}[c]{@{}l@{}}$+\mathcal L_{\mathrm{roll}}+\mathcal L_{\mathrm{read}}$\\ (\textbf{QL-VLM})\end{tabular}
& 0.05 & 0.05 & 1 & \textbf{90.43} & \textbf{93.78} \\
\bottomrule
\end{tabular}
\end{table}

\begin{figure}[htbp]
    \centering
    \includegraphics[scale=.45]{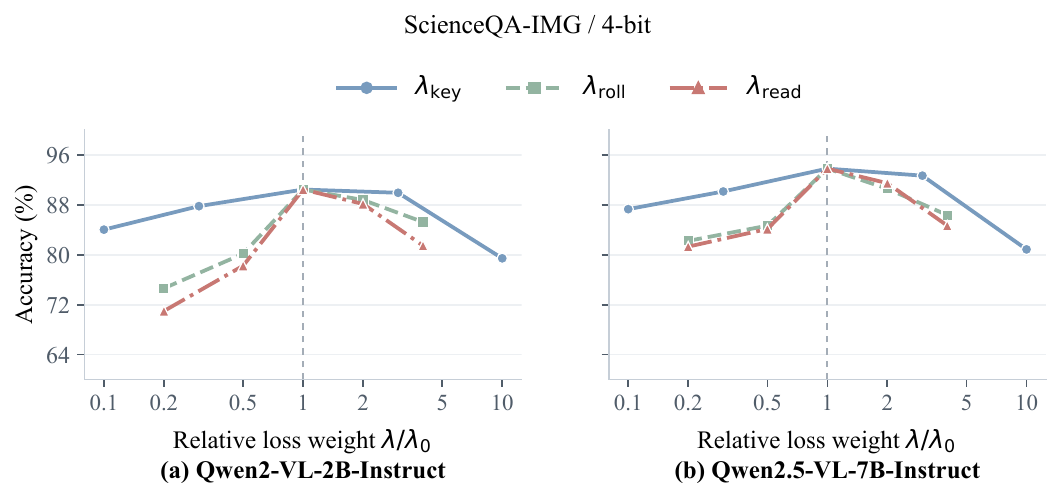}
    \caption{Loss-weight sensitivity on ScienceQA-IMG at 4-bit. Each curve varies one coefficient while fixing the others. The horizontal axis shows the ratio to its default: $\lambda_{\mathrm{key}}=0.001$, $\lambda_{\mathrm{roll}}=0.05$, and $\lambda_{\mathrm{read}}=0.05$. Dashed lines denote default.}
\label{fig:loss_weights}
    \label{fig:loss_wight}
\end{figure}

\noindent\textbf{Hyper-parameter selection.}
Fig.\,\ref{fig:loss_weights} examines the three auxiliary loss weights independently. Both models perform best at default. Tripling the key-loss weight reduces accuracy by at most 1.13\%, whereas a tenfold increase causes an 11.01\%-12.93\% drop. Reducing the readout weight to 1/5 of its default also lowers accuracy by 12.49\%-19.45\%. These trends suggest that insufficient reasoning supervision and excessive key reconstruction regularization can both impair performance, supporting a balanced weighting of reconstruction and predictive alignment.

\noindent\textbf{Latent budget and compression trade-off.}
Tab.\,\ref{tab:latent_budget} shows that insufficient latent budgets degrade both full-precision and quantized reasoning, with QL-VLM improving substantially as $T$ increases from 4 to 6. Beyond 6 thoughts, accuracy gains diminish: increasing $T$ from 6 to 12 improves QL-VLM by only 1.22\% , while reducing compression ratio from 4.75$\times$ to 3.15$\times$. QL-VLM outperforms the baseline at every budget. We therefore adopt 6 thoughts as a practical balance between accuracy and reasoning compression.

\begin{table}[htbp]
\centering
\small
\caption{Latent-budget trade-off on ScienceQA with Qwen2-VL-2B-Instruct. Steps and CR denote decoding length and compression ratio of QL-VLM relative to explicit CoT.}
\label{tab:latent_budget}
\begin{tabular}{cccccc}
\toprule
$T$ & Latent CoT & Baseline & QL-VLM & Steps & CR \\
\midrule
2  & 63.21 & 18.17 & 45.15 & 6.0 + 2 & 7.04 \\
4  & 68.12 & 22.46 & 53.61 & 6.03 + 4 & 5.61 \\
6  & 79.97 & 50.10 & 90.43 & 5.86 + 6 & 4.75 \\
8  & 81.26 & 55.23 & 91.01 & 5.93 + 8 & 4.04 \\
12 & 82.94 & 56.79 & 91.65 & 5.86 + 12 & 3.15 \\
\bottomrule
\end{tabular}
\end{table}

\section{Conclusion}
We introduced QL-VLM for efficient latent reasoning, addressing both reasoning length and the cost of low-precision execution. Our diagnosis identifies key sensitivity and direction-dependent predictive effects of continuous-thought errors. Therefore, DGQ preserves rotary-key structure through nonuniform quantization, while DPD aligns continuous thoughts, latent-query readouts, and predictions with a full-precision teacher. 
% 
% Experiments across VLMs and tasks show improved low-bit accuracy and cross-dataset generalization, while retaining compact reasoning, reducing theoretical model storage, and achieving faster measured inference.
Experiments across VLMs and tasks show improved low-bit accuracy and cross-dataset generalization, alongside compact reasoning, lower theoretical storage, and less measured inference time.

% \section*{LLM Usage Statement}
% We used OpenAI Codex to assist with manuscript drafting and language editing, code assistance including figure and table preparation. The authors are responsible for the scientific claims, experimental results, references, and final manuscript.

{
    \small
    \bibliographystyle{ieeenat_fullname}
    \bibliography{main}
}

\end{document}